\pdfoutput=1

\documentclass[11pt]{article}

\usepackage[final]{acl}

\usepackage{times}
\usepackage{latexsym}

\usepackage{makecell} 

\usepackage{amssymb}
\usepackage{siunitx}

\usepackage[T1]{fontenc}

\usepackage[utf8]{inputenc}

\usepackage{microtype}

\usepackage{inconsolata}

\usepackage{graphicx}

\usepackage{multirow}
\usepackage{colortbl}
\usepackage{bbm}

\usepackage{xcolor}
\usepackage{nicematrix}

\usepackage{booktabs}
\usepackage{multirow}
\usepackage{adjustbox}

\title{Assessing Y-Axis Influence: Bias in Multimodal Language Models on Chart-to-Table Translation}

\author{Seok Hwan Song, Azher Ahmed Efat, Wallapak Tavanapong \\ Department of Computer Science, Iowa State University, Ames, Iowa, USA \\ \texttt{\{song92, efat, tavanapo\}@iastate.edu}}

\begin{document}

\maketitle
\begin{abstract}

Chart-to-table translation converts chart images into structured tabular data. Accurate translation is crucial for Multimodal Language Model (MLM) to answer complex queries. We observe imbalances in the number of images across different aspects of the y-axis information in public chart datasets. Such imbalances can introduce unintended biases, causing uneven MLM performance. Previous works have not systematically examined these biases. To address this gap, we propose a new framework, \textbf{FairChart2Table}, for analyzing y-axis-related bias on five state-of-the-art models.

\textbf{Key Findings:} (1) There are significant y-axis biases related to the digit length of the major tick values, the number of major ticks, the range of values, and the tick value format (e.g., abbreviation or scientific format). (2) The number of legends/entities in chart images impacts MLM performance. (3) Prompting MLM with y-axis information can significantly enhance the performance for some MLMs. 
\end{abstract}

\section{Introduction}

Multimodal Language Models (MLMs) have been evaluated on diverse tasks in chart image understanding, such as chart image summarization \citep{islam-etal-2024-large}, question answering (QA) \citep{chartqa}, fact checking \citep{akhtar-etal-2023-reading}, and chart-to-table translation---translation of data in chart images to corresponding tabular data \citep{deplot}. 

Biases in MLM performance on QA have been studied. These biases include chart complexity measured by the number of entities (data series), color scheme, font size, grid lines, legend position, scale of the figure, logarithmic scale in the y-axis ticks, and the x-axis tick orientation \citep{unraveling}, along with 
chart image resolutions \citep{resolution}.

\begin{table}[!thbp]
\centering
\small
\begin{NiceTabular}{l|c|c}
\toprule
\textbf{Dataset}
  & \makecell{\textbf{Digit Length}\\\textbf{Range}}
  & \makecell{\textbf{CV}\\(Std./Mean$^*$)} \\
\toprule
PlotQA
  & 0–16
  & \makecell{\textbf{1.72}\\(15040.9/8769.6)} \\
ChartQA
  & 0–11
  & \makecell{\textbf{1.40}\\(2441.0/1738.1)} \\
\midrule
\makecell[l]{FairChart2Table\\(Ours - Part A)}
  & 0–16
  & \makecell{\textbf{0.00}\\(0.0/180.0)} \\
\bottomrule
\end{NiceTabular}
\caption{Imbalance in datasets by digit length; CV stands for coefficient of variation \citep{cv}. $^*$Mean and Std. are in terms of \#images per digit length.}
\label{tbl:cvstat}
\end{table}

This paper focuses on unexplored biases related to the y-axis information in chart-to-table translation tasks. Accurate chart-to-table translation enables more accurate reasoning than direct image-based QA \citep{simplot}, and improves interpretability in complex reasoning with the extracted tabular data. The y-axis has different information important for accurate value extraction, such as the minimum and maximum values on the y-axis, the number of major and minor ticks, and the tick value format---the format of the number placed next to the tick marks, such as a scientific format, numbers with commas and abbreviations like {\bf K}, {\bf M}, {\bf B}, {\bf T} for a thousand, a million, a billion, and a trillion, respectively. 

To observe a potential bias related to the y-axis, we analyzed two well-known chart datasets, PlotQA~\citep{plotqa} and ChartQA~\citep{chartqa}.
We characterize the y-axis maximum value by the number of digits in its integer part, referred to hereafter as {\em digit length}. For example, a number in the range of (-1, 1) has zero digit length; numbers in the range of [1,10) or [-1, -10) have the digit length of 1. The longer the digit length, the larger the positive or negative value. Table \ref{tbl:cvstat} shows the imbalance in the number of images across digit lengths in the existing datasets and our proposed dataset without such biases (CV=0).

This imbalance, along with other imbalances related to the y-axis information, can lead to unintended biases, ultimately impacting MLM performance in chart-related tasks. Consequently, models may excel on chart images with specific y-axis characteristics but struggle to generalize to others. 

{\em To our knowledge, the influence of the y-axis information on MLM performance for chart-to-table translation has not been systematically investigated, and no benchmark for the evaluation exists.} 
 We propose a new {\bf Bias Controlled Chart-to-Table (FairChart2Table) Framework}. The framework includes (1) dataset generation methods designed to eliminate confounding biases on the y-axis and (2) new performance metrics, considering visually noticeable errors more seriously than small fluctuations. {\em To quantify these visible errors, we define Tick-Based Error (TBE) as the ratio of the absolute difference between the ground truth and the predicted value to the minor tick interval.} Using our framework, we investigate the following research questions for chart-to-table translation.

\begin{enumerate}
\item (RQ1) How does the y-axis information affect MLM performance? 
\item (RQ2) What other factors of chart images (e.g., chart styles, crossing lines) impact MLM performance?
\item (RQ3) Does prompting MLMs with explicit y-axis information help improve MLM performance?

\end{enumerate}

Our contributions are summarized below.

{\bf Contribution \#1:} The aforementioned research questions that have not been explored.

{\bf Contribution \#2:} The FairChart2Table framework, consisting of the methodology to generate the benchmark dataset for the above research questions and the new TBE-based performance metrics. We will share our code and dataset publicly\footnote{Code and benchmark: \url{https://github.com/NRT-D4/FairChart2Table}}. The benchmark can be used to test biases in other MLMs.

{\bf Contribution \#3:} New key findings from evaluations of five models on chart-to-table translation. These models include two closed-source, one open-source, and three open-source fully supervised models.  Some key findings are as follows. (1) Y-axis related biases exist across multiple configurations, including the number of major ticks, the numerical range of the axis, and the tick value format. These factors can significantly influence model performance. (2) Models tend to get confused about the position of numerical values on the same x-axis tick when the number of entities, corresponding to the number of distinct elements in the chart legend, increases. In contrast, the chart type appears to have a minor influence. (3) Prompting MLMs with the y-axis tick values for chart-to-table translation can enhance model performance.

\begin{figure*}[h!tb] {
\centering
\includegraphics[width=1\textwidth]{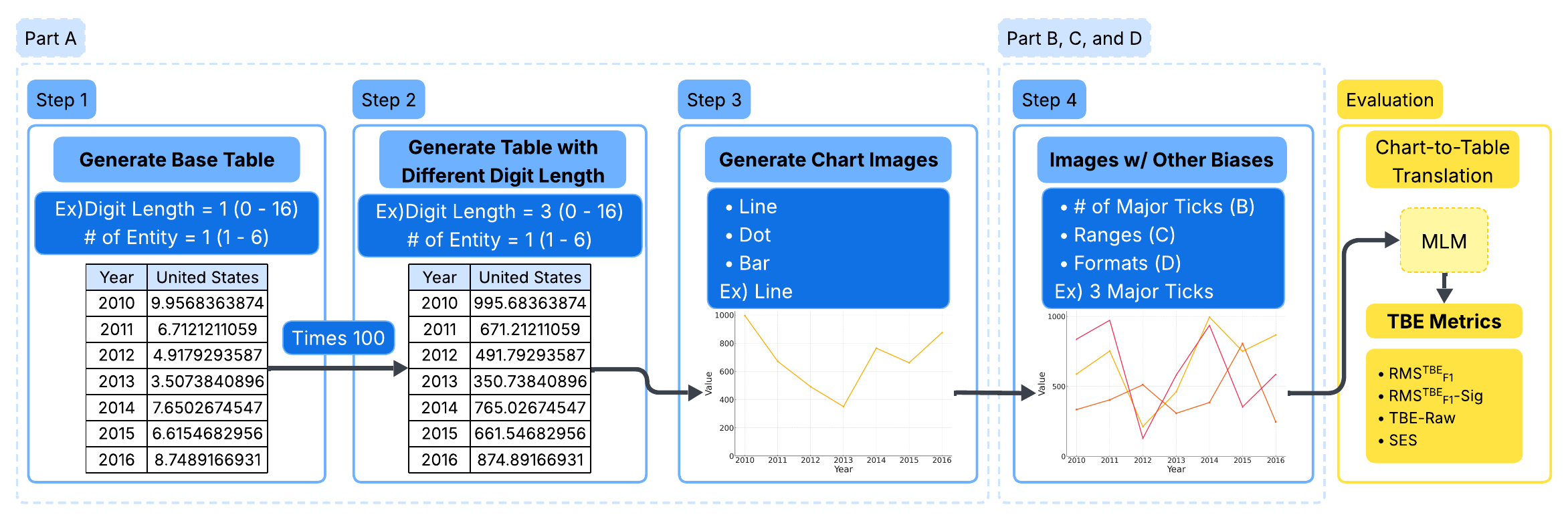} 
\caption{FairChart2Table Framework. Step 1 generates 10 base tables for each entity count (1-6) with a digit length of 1. Step 2 derives 1,020 tables by scaling the values to the target digit length (0-16) using powers of ten. Step 3 creates 3,060 chart images representing three chart types based on the generated tables. Step 4 chooses only the tables with three entities from Part A, and produces additional images by varying the number of major ticks, ranges, and formats. Last, evaluate MLMs for chart-to-table translation using our TBE-based metrics.}
\label{framework}}
\end{figure*}

\section{Related Work \label{related_work}}

\subsection{Chart Benchmarks}


{\bf Chart Datasets:} DVQA \citep{dvqa} and FigureQA \citep{figureqa} are among the earliest datasets introduced for chart-based question answering. 
PlotQA \citep{plotqa} and ChartQA \citep{chartqa} are widely used for chart-based question answering. 
PlotQA has synthetic plots created from real-world data, and the questions were generated using templates. ChartQA consists of human-written questions as well as automatically generated questions from human-written summaries. LEAF-QA \citep{leafqa} is a large dataset with 2 million question-answer pairs. Recent work includes datasets on multiple chart images \citep{multichartqa, mmc, chartToText}.

 {\bf Performance Metrics for Chart-to-Table Translation:} Relative Mapping Similarity F1 (RMS$_{F1}$), proposed by \citet{deplot} is widely used~\citep{tinychart, simplot, unichart, chartassist}. This metric assesses the alignment between the predicted and ground truth tables, considering both textual and numeric elements and the table structure (column and row headers). RMS$_{F1}$ calculates the F1 score using the combined distance function, incorporating the normalized Levenshtein distance for textual mismatch, and the relative numeric distance for numerical values. Minimal cost matching is used to match the values between the target and predicted tables.

{\bf Models:} Recent models are categorized into three groups. 1) Specialized models such as DePlot \citep{deplot} and Simplot \citep{simplot} were designed for chart-to-table translation by leveraging encoder-decoder architectures tailored to visual inputs. 2) Matcha \citep{matcha}, UniChart \citep{unichart}, and TinyChart \citep{tinychart} are supervised models trained on general chart understanding tasks. 3) General-purpose MLMs like GPT-based multimodal models such as GPT-4o offering promising results on chart-related tasks \citep{multichartqa,islam-etal-2024-large,unraveling}. Instruction tuning was shown to enhance performance on chart reasoning tasks \citep{chartgemma, chartinstruct}. \citet{color} pretrained existing models on simple tasks, structural and visual knowledge, data extraction, and numeric question answering.

\subsection{Biases in MLMs for Chart-Based Tasks}

\begin{table}[!thbp]
\centering
\small
\begin{NiceTabular}{l|c}
\toprule
\textbf{Aspect} & \textbf{Factor Manipulation} \\
\toprule
Complexity & Single entity and $\geq$2 entities \\
Color scheme &  palettes, gradients, and hues \\
Font size & Small and Big \\
Grid lines & Presence / density \\
Legend position & Placement \\
Figure scale & Overall size \\
Tick scale (log) & Linear and log \\
Tick orientation & Horizontal and angled/vertical \\
\midrule
Image resolution & pixel count \\
\bottomrule
\end{NiceTabular}
\caption{Existing work on bias analysis of MLM performance~{\citep{unraveling,resolution}} \label{bias-comp}}
\end{table}

Table~\ref{bias-comp} summarizes existing work on bias analysis. Bias from digit length was not considered. Our work also investigates the impact of varying the number of entities from 1 to 6, more than the number studied in \citep{unraveling}. Unlike the work by \citet{unraveling}, which evaluates model performance solely on a log scale in ticks, we consider three other types of tick value number format (comma, scientific notation, and abbreviation types). 

\section{Proposed Bias Controlled Chart-to-table (FairChart2Table) Framework}

Current datasets contain diverse components, such as chart types, styles, and content, which can introduce various forms of bias. As a result, it is challenging to use them to effectively identify and evaluate biases in the chart-to-table translation task. Additionally, existing evaluation metrics are insufficient to capture easily noticeable visual errors, thereby limiting the reliability of current assessments. To address these issues, we propose \textbf{Bias Controlled Chart-to-Table (FairChart2Table) Framework} with a new method to generate a new dataset free from bias related to the y-axis information and new evaluation metrics.
The proposed framework enables investigations of our research questions and can be used to evaluate the y-axis biases in other MLMs.

\subsection{Benchmark Construction}

Figure~\ref{framework} illustrates the process to generate our benchmark. To avoid other unintended biases, we generated all chart images using the Bokeh plotting library as in PlotQA \citep{plotqa}, one of the most widely adopted datasets for training state-of-the-art models \citep{deplot, unichart, tinychart}. Additionally, chart elements, including titles, units, labels, font styles, font sizes, legend locations, x-axis information, and color combinations, are standardized. 

The dataset has four subsets (A-D), each focusing on a distinct y-axis-related bias: digit length, number of entities, number of major ticks, value ranges, and tick value formats. Each dataset is designed to isolate a single bias, ensuring no confounding factors.

\textbf{Part A for the evaluation of digit length bias:} We generated 10 tables for each entity count from 1 to 6, using values within a single-digit length, 0 to 10. These values in the tables are scaled by powers of ten to the target digit length from 0 to 16. From these 1,020 (10*6*17) tables, we generated one line, one dot, and one bar chart. The y-axis has 6 major ticks, with the tick values in plain numerical format, and the intersection of the x-axis and y-axis at the origin. This dataset enables a fair evaluation of model performance across different digit lengths and entity counts. Part A has a total of 3,060 chart images.


\textbf{Parts B-D:} To eliminate the number of entities as a confounding factor, we chose only the tables with three entities from Part A, resulting in 170 tables to use as the base for generating the remaining parts. The specific details for each are below.

{\bf Part B for the evaluation of major tick bias} has 170 chart images with 3 major ticks and 170 chart images with 11 major ticks for performance comparison with the chart images with 6 major ticks and three entities from Part A.  

{\bf Part C for the evaluation of y-axis range bias} has 3*170 chart images with three different ranges of y-axis values, varying from those of the 170 tables in Part A as follows. 
\begin{itemize}
\item Positive minimum tick value (Pos): All values are shifted upward by adding three times the major tick interval, ensuring the minimum tick value is positive. The corresponding chart has a positive displaced y-axis origin.
\item Negative minimum tick value (Neg): All values are shifted downward by subtracting three times the major tick interval, ensuring the minimum tick value is negative. The corresponding chart has a negative displaced y-axis origin.
\item Extended range (Ext): The data values remain unchanged, but the maximum tick value is double, visually pushing the data values to the bottom of the chart.

\end{itemize}

We do not characterize the minimum tick values by the digit length since we did not observe a performance impact due to the digit length of the negative values.

\textbf{Part D for the evaluation of number format bias} has chart images with three variants of the y-axis label formats. Numbers can be represented with commas (e.g., 7,000), abbreviations (K, M, B, and T), and a scientific format (e.g., 7.00e + 6).

The total of chart images for this dataset is 7,140 images (3060 + (2*170 + 2*3*170)*3). Appendix~\ref{app:image} shows chart image examples.

\begin{figure}[!htb] {
\centering
\includegraphics[width=0.49\textwidth]{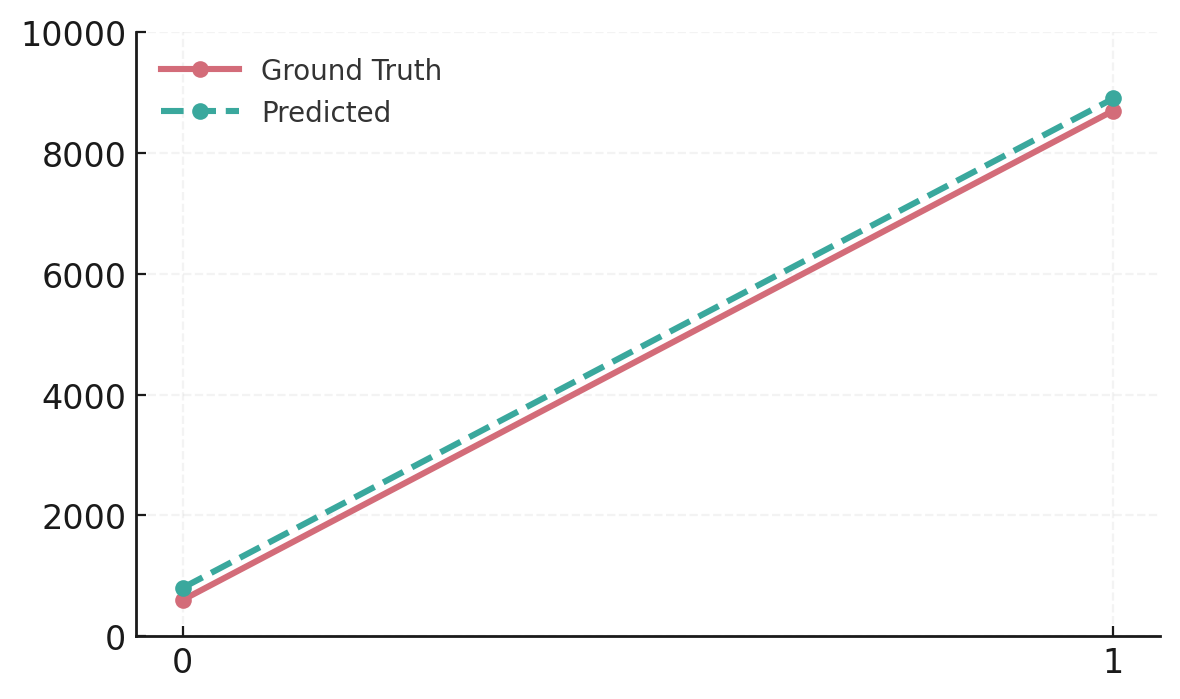} 
\caption{Ground truth and predicted values differ by the same amount, 200 points, at both $x = 0$ and $x = 1$. For these two points, the distance measured by RMS$^{TBE}_{F1}$ is 0.5 in both cases, while RMS$_{F1}$ yields 0.33 at $x=0$ and 0.02 at $x=1$. Visually, RMS$^{TBE}_{F1}$ reflects the equal difference between the predicted and the ground truth values better.}
\label{tbe_chart}}
\end{figure}

\subsection{Evaluation Metrics}

RMS$_{F1}$ \citep{deplot} is a widely used metric for table-based evaluation. It first aligns the predicted and ground truth tables using its own matching procedure and then scores the aligned cells with Eq.~\ref{rms} as the distance. 

\begin{equation} D_{\text{RMS}}(g, p) = min \left(  1,\frac{\lvert g - p \rvert}{\lvert g \rvert} \right), \label{rms} \end{equation}
where $g$ and $p$ reflect the ground truth and predicted values of aligned cells.
However, the distance function does not capture visual differences well.
As illustrated in Figure~\ref{tbe_chart}, visually equivalent differences can receive different scores under $D_{\text{RMS}}$, leading to inconsistent scoring for the chart-to-table task.

We introduce and investigate several \textbf{Tick-Based Error (TBE)} metrics. TBE in Eq.~\ref{tbed} normalizes the absolute difference between the ground truth $g$ and the predicted value $p$ by a fixed amount $t>0$ that estimates the minor tick interval. As Figure~\ref{tbe_chart} shows, the distance of RMS$^{TBE}_{F1}$ gives visually more accurate scores. Our rationale is that an error larger than a minor tick interval is visually noticeable on a chart. Our implementation uses one-fifth of the major tick interval of an input chart to estimate the minor tick interval to ensure that TBE can be computed for all charts, including those without minor ticks.

\subsubsection{Proposed RMS$^{TBE}_{F1}$ Metric}

The calculation of RMS$^{TBE}_{F1}$ score follows that of the RMS$_{F1}$ score~\citep{deplot} except for two changes. (1) We use the TBE distance in Eq.~\ref{tbed} for the numerical error.
\begin{equation} D_{\text{TBE}}(g, p) = min \left(  1,\frac{\lvert g - p \rvert}{t} \right) \label{tbed} \end{equation}

(2) Once the column and row headers of the truth and predicted table from chart-to-table translation are matched, no further penalties for headers are applied to the associated value cells. As \citet{simplot} noted, Levenshtein distances used in RMS$_{F1}$ for column and row headers may not be close for synonym words. Moreover, MLMs sometimes incorporate information from the titles and labels of a chart into the predicted headers, which do not alter the meaning of the headers but decrease the RMS$_{F1}$ score. {Although we still use the matching procedure in RMS, using \textit{Normalized Levenshtein Distance} to match headers between the truth and the predicted tables, we do not repeatedly apply it across cells, reducing over-penalization and improving fairness in the overall error aggregation.}

Since our analysis focuses on biases with numerical information, we exclude the header similarity scores from the main results. For completeness, Appendix~\ref{app:rms-results} presents both RMS$_{F1}$ and RMS$^{TBE}_{F1}$ to demonstrate over-penalization due to the table header.
RMS$^{TBE}_{F1}$ values are in the interval [0,1]. The higher the values, the closer the predicted values are to the ground truth values.
Appendix ~\ref{sec:appendix_rms} shows the equation for RMS$^{TBE}_{F1}$.

\subsubsection{Proposed RMS$^{TBE}_{F1}$-Sig and TBE-Raw Metrics}

These metrics employ the same mapping procedures as RMS$^{TBE}_{F1}$ but differ in their choice of distance function and aggregation scheme. Specifically, RMS$^{TBE}_{F1}$ employs $D_{TBE}$ for the distance and aggregates through the harmonic mean ($F1$).
The distance used in RMS$^{TBE}_{F1}$-Sig shown in Eq.~\ref{tbed-seg} considers only \textit{significant} deviations---those with the absolute error greater than the value of $t$, to avoid the dilution effect in tables with many data points. 
\begin{equation} 
D_{\text{TBE-Sig}}(g, p) = 
\mathbf{1}\!\left\{\frac{\lvert g - p \rvert}{t}\geq 1 \right\},
\label{tbed-seg} 
\end{equation}
where $\mathbf{1}\left\{c\right\}$ is an indicator function that returns 1 when the condition $c$ is true; otherwise 0.
The metric RMS$^{TBE}_{F1}$-Sig considers only the cells with significant deviations in the calculation.

The TBE-Raw metric utilizes Eq.~\ref{tbe-raw} to calculate the distance between individual aligned cells in the predicted and truth tables. This distance function does not suppress any error magnitudes as in RMS$^{TBE}_{F1}$. The metric TBE-Raw is the mean of all cell's $D_{\text{TBE-Raw}}$ from the tables. Therefore, the values are in the range of [0,$\infty$).

\begin{equation} 
D_{\text{TBE-Raw}}(g, p) = \frac{\lvert g - p \rvert}{t} \label{tbe-raw} 
\end{equation}

\subsubsection{Swapping Error Score (SES)}

We introduce \textbf{Swapping Error Score (SES)} shown in Eq.~\ref{eq:ses} to measure errors commonly observed when numbers are correctly extracted, but the entities they belong to are swapped. This situation often occurs when MLMs process a chart image with intersecting lines. At a given x-axis value, the extracted y-value of one line (entity) is assigned to an entity of a crossing line and vice versa. 
\begin{equation}
\text{SES} = \text{RNSS$^{TBE}_{F1}$} - \text{RMS$^{TBE}_{F1}$,}
\label{eq:ses}
\end{equation}
where RNSS$^{TBE}_{F1}$ measures the similarity between the values in the ground truth and the predicted tables. To calculate RNSS$^{TBE}_{F1}$, we use a minimal cost matching algorithm to minimize relative errors across all value pairs, regardless of the order, as described in \citep{deplot}. However, we make one difference: utilizing D$^{TBE}$ for numerical error calculations and the harmonic mean. We subtract RMS$^{TBE}_{F1}$ to isolate the impact of swapped numbers, offering a clearer view of structural errors beyond the accuracy of the raw value. SES values are in the range of [-1, 1]. 

None of these metrics, including the original RMS$_{F1}$, captures the differences in the chart titles. 

\begin{figure}[h!tb] {
\centering
\includegraphics[width=0.49\textwidth]{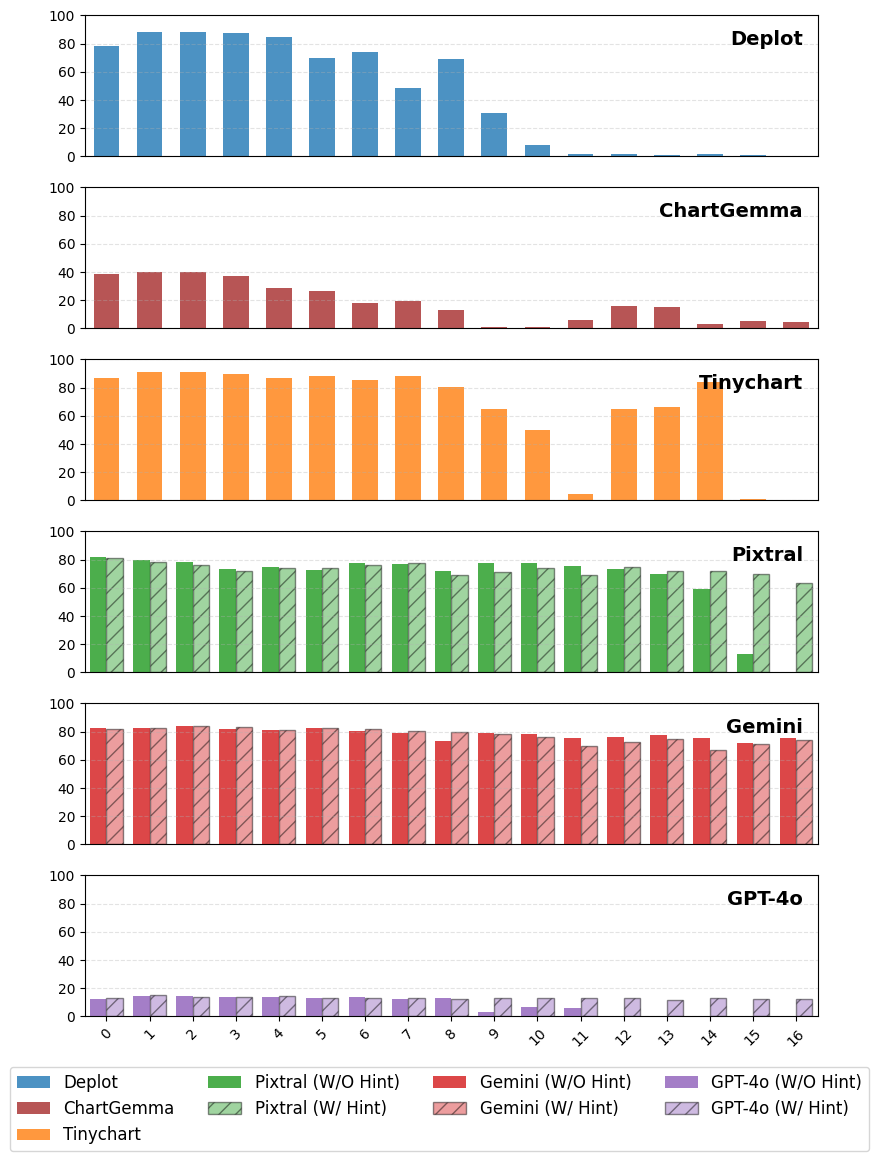} 
\caption{RMS$^{TBE}_{F1}$ performance for bias evaluation by digit length (X-axis); ``W/Hint'' indicates when the y-axis major tick values were used as part of the prompts to MLMs. 
\label{scales}}}
\end{figure}

\begin{table}[t]
\centering
\small
\begingroup
\setlength{\tabcolsep}{3pt}
\renewcommand{\arraystretch}{1.05}
\sisetup{group-digits = false, table-number-alignment = center}

\begin{tabular}{@{}c|S[table-format=2.2]|S[table-format=2.2]|S[table-format=2.2]|S[table-format=2.2]|S[table-format=2.2]|S[table-format=5.2]@{}}
\toprule
\textbf{DL} &
\multicolumn{1}{c|}{\textbf{DePlot}} &
\multicolumn{1}{c|}{\makecell{\textbf{Chart}\\\textbf{Gemma}}} &
\multicolumn{1}{c|}{\makecell{\textbf{Tiny}\\\textbf{Chart}}} &
\multicolumn{1}{c|}{\textbf{Pixtral}} &
\multicolumn{1}{c|}{\textbf{Gemini}} &
\multicolumn{1}{c}{\textbf{GPT-4o}} \\
\midrule
0  & 3.06 & 4.01  & 0.26  & 0.38 & 0.39 & 3.41 \\
1  & 0.53 & 3.79   & 0.16  & 0.47 & 0.37 & 3.39 \\
2  & 0.58 & 3.63 & 0.16  & 0.49 & 0.31 & 3.49 \\
3  & 0.67 & 3.90 & 0.17  & 0.57 & 0.43 & 3.50 \\
4  & 1.07 & 5.70 & 0.42  & 0.55 & 0.45 & 3.44 \\
5  & 3.03 & 6.85 & 0.35  & 1.23 & 0.38 & 3.63 \\
6  & 2.44 & 8.95 & 0.54  & 0.72 & 0.47 & 3.68 \\
7  & 5.93 & 8.21 & 0.38  & 0.86 & 0.50 & 4.44 \\
8  & 3.06 & 9.67 & 8.61  & 1.55 & 1.21 & 4.00 \\
9  & 8.33 & 12.12 & 27.18 & 1.20 & 0.60 & 10.47 \\
10 & 11.19 & 12.52 & 30.22 & 0.61 & 0.56 & 13.12 \\
11 & 12.08 & 15.38 & 82.27 & 0.78 & 0.70 & 8.61 \\
12 & 12.05 & 54.92 & 15.47 & 0.71 & 0.74 & 12.39 \\
13 & 12.18 & 12.63 & 12.45 & 0.78 & 0.64 & 13.14 \\
14 & 12.04 & 20.36 & 0.84  & 3.25 & 0.68 & 13.50 \\
15 & 12.62 & 16.01 & 12.21 & 10.25 & 4.12 & 62843.85 \\
16 & 13.12 & 11.22 & 13.08 & 12.90 & 1.57 & 13.63 \\
\bottomrule
\end{tabular}
\caption{TBE-Raw performance per digit length (DL). The higher the values, the further the predicted values are from the ground truth values.}
\label{raw}
\endgroup
\end{table}

\subsection{Models and Experiments}

We selected TinyChart \citep{tinychart} to represent fully supervised models for tasks in chart image understanding, since it performs the best on chart-to-table translation among SOTA models \citep{tinychart}. We additionally included ChartGemma \citep{chartgemma}, since it is, like TinyChart, a model trained on multiple chart understanding tasks. DePlot \citep{deplot} was selected to represent fully supervised models only for chart-to-table translation, since it performs better than UniChart overall. Simplot was not compared since it requires additional fine-tuning for different chart styles \citep{simplot}. For closed-source MLMs, we chose GPT-4o (gpt-4o-2024-05-13)~\citep{4o} and Gemini-2.0-flash, and chose Pixtral-12B-2409 \citep{pixtral} to represent an open-source MLM for repeatability of experiments. The default configurations used greedy decoding (i.e., temperature = 0). For each reasoning problem and configuration, a single API call was made to an MLM to facilitate paired statistical tests. The experiments were conducted in late September 2025.

\subsection{Statistical Test}

Hereafter, we present all metric values as percentages, those ranging from 0 to 1 (RMS$^{TBE}_{F1}$ and RMS$^{TBE}_{F1}$-Sig), and those ranging from -1 to 1 (SES). We conducted paired two-sided Wilcoxon signed-rank tests comparing the performances of a given model on the same charts under each condition against the baseline, treating each chart as a paired observation. We employed rank-based tests for the metrics with the same bounded ranges, ensuring robustness due to departures from normality and the presence of many outliers. 

\begin{table*}[ht]
\centering
\small
\begin{tabular}{l|c|cc|ccc|ccc}
\toprule
\textbf{Model} 
& \textbf{Base Conf.} 
& \multicolumn{2}{c|}{\textbf{\#Major ticks}} 
& \multicolumn{3}{c|}{\textbf{Range}} 
& \multicolumn{3}{c}{\textbf{Format}} \\
\cmidrule(lr){2-2} \cmidrule(lr){3-4} \cmidrule(lr){5-7} \cmidrule(lr){8-10}
& 
& 3 & 11 
& Pos & Neg & Ext 
& Comma & Sci.\ & Abbr.\ \\
\toprule

\multicolumn{10}{l}{\textbf{Open Source}} \\
DePlot 
& 45.48 
& \cellcolor{cyan}27.88 & \cellcolor{pink}47.94 
& \cellcolor{pink}46.81 & \cellcolor{cyan}29.19 & \cellcolor{cyan}9.07 
& \cellcolor{pink}63.79 & \cellcolor{cyan}17.76 & \cellcolor{cyan}21.61 \\

ChartGemma 
& 19.34 
& \cellcolor{cyan}6.68 & 21.45 
& \cellcolor{cyan}9.25 & \cellcolor{cyan}9.92 & \cellcolor{cyan}3.43 
& 20.77 & \cellcolor{cyan}1.38 & \cellcolor{cyan}9.92 \\

TinyChart 
& 72.56
& \cellcolor{cyan}33.24 & \cellcolor{cyan}67.45
& \cellcolor{cyan}25.17 & \cellcolor{cyan}7.00 & \cellcolor{cyan}45.86
& \cellcolor{cyan}37.11 & 80.65 & \cellcolor{cyan}26.29 \\
Pixtral 
& 67.88 
& \cellcolor{cyan}29.16 & \cellcolor{pink}79.11 
& \cellcolor{cyan}42.43 & \cellcolor{cyan}11.15 & \cellcolor{cyan}31.63 
& \cellcolor{cyan}58.92 & 71.89 & \cellcolor{cyan}32.14 \\
\midrule

\multicolumn{10}{l}{\textbf{Closed Source}} \\
Gemini 
& 83.01 
& \cellcolor{cyan}41.77 & \cellcolor{pink}85.58 
& \cellcolor{cyan}80.22 & \cellcolor{cyan}71.68 & \cellcolor{cyan}44.32 
& 83.26 & \cellcolor{cyan}82.62 & \cellcolor{cyan}46.54 \\
GPT-4o 
& 8.84 
& \cellcolor{cyan}6.50 & \cellcolor{pink}10.75 
& \cellcolor{pink}10.44 & \cellcolor{pink}11.91 & \cellcolor{cyan}5.47 
& \cellcolor{pink}14.12 & \cellcolor{pink}13.44 & \cellcolor{cyan}6.19 \\
\midrule

\multicolumn{10}{l}{\textbf{Prompting with major tick values (Ours)}} \\
Pixtral 
& 76.62 
& \cellcolor{cyan}44.14 & \cellcolor{pink}85.13 
& \cellcolor{cyan}49.47 & \cellcolor{cyan}14.86 & \cellcolor{cyan}37.01 
& \cellcolor{cyan}73.97 & \cellcolor{cyan}72.83 & \cellcolor{cyan}58.74 \\
Gemini 
& 83.77
& \cellcolor{cyan}42.12 & \cellcolor{pink}85.61 
& \cellcolor{cyan}79.71 & \cellcolor{cyan}74.42 & \cellcolor{cyan}49.91 
& 83.80 & \cellcolor{cyan}82.90 & \cellcolor{cyan}61.32 \\
GPT-4o  
& 13.22 
& \cellcolor{cyan}11.00 & \cellcolor{pink}16.31 
& 12.90 & 13.38 & \cellcolor{cyan}7.01 
& \cellcolor{cyan}12.33 & \cellcolor{cyan}12.29 & \cellcolor{cyan}11.10 \\
\bottomrule
\end{tabular}

\caption{RMS$^{TBE}_{F1}$ performance for the base configuration (with only 3 entities, 6 major ticks, the x-axis and y-axis intersecting at the origin, and the plain numeric tick value format) and others. Statistically significant cases were highlighted. \colorbox{cyan}{Cyan} and \colorbox{pink}{pink} indicate that the base configuration is better or worse than the compared configurations, respectively.}
\label{everything}
\end{table*}

\begin{figure*}[h!tb] {
\centering
\includegraphics[width=1\textwidth]{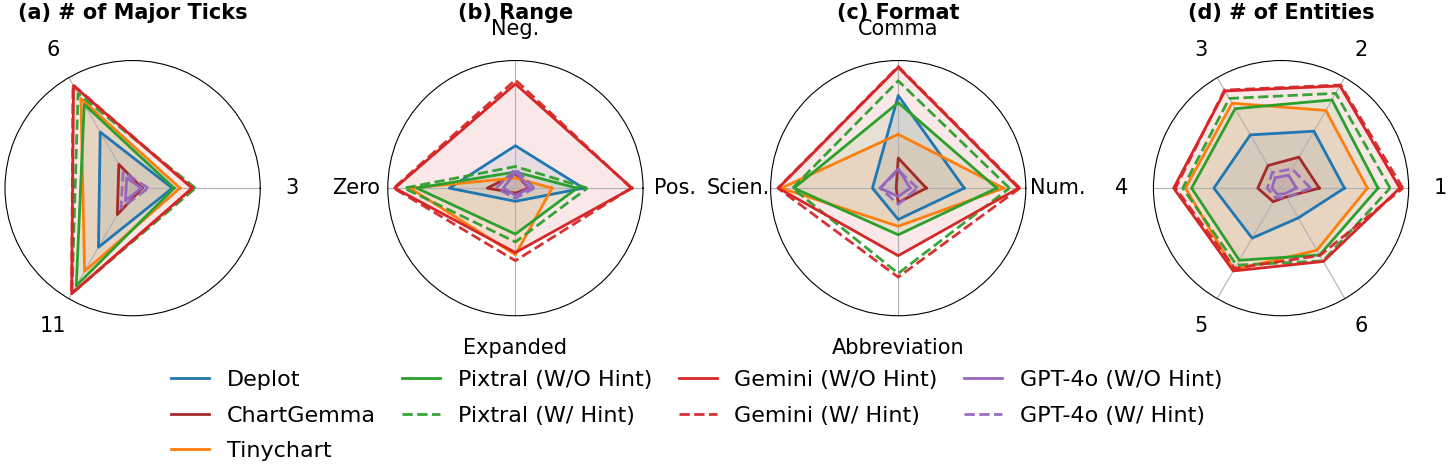} 
\caption{RMS$^{TBE}_{F1}$ performance for bias evaluation associated with the number of major ticks, y-axis value range, tick number format, and the number of entities
\label{spider}}}
\end{figure*}

\section{Experimental Results}

\subsection{RQ1: How does the y-axis information affect MLM performance in chart-to-table translation? \label{RQ1}}

\noindent{\bf Biases by digit length (measured by RMS$^{TBE}_{F1}$):} 
Figure~\ref{scales} shows performance variation by digit length using our FairChart2Table dataset Part A. There are 60 tables per digit length.
DePlot's performance generally declines as the digit length increases, except for the case of digit length 0.
The performance at digit length 1 is the best, which is also the case for charts with the maximum y-axis value in the range [1, 10). ChartGemma exhibits relatively low overall performance among the compared models, with a general downward trend as digit length increases up to 10, followed by a slight recovery at larger digit lengths. TinyChart's performance generally decreases with increasing digit lengths. Extremely low performance occurs at digit lengths 11, 15, and 16.
Pixtral exhibits a similar trend to TinyChart, with the decrease being largest at digit lengths of 15 and 16.
Compared to the other models, Gemini is the least biased and achieves the highest performance with digit lengths ranging from 6 to 16.  However, its performance still decreases with increasing digit lengths.
GPT-4o performs the worst among the compared models and still exhibits bias by digit lengths.

\noindent{\bf To what extent models are affected by digit length (measured by TBE-Raw):} High values of TBE-Raw indicate large errors.
Table~\ref{raw} shows that at some digit lengths, there are unusually large errors. At digit length 15, GPT-4o has an extremely large error, differing from that at digit length 16, but the corresponding RMS$^{TBE}_{F1}$ in Figure~\ref{scales} are both under 15\%.  DePlot's raw error at digit length 0 is relatively larger than those at digit lengths 1-4.
Similar patterns are observed with ChartGemma at digit length 12, TinyChart at digit lengths 9 and 10, and with Gemini at digit length 15.

\begin{table*}[htbp]
\centering
\scriptsize 
\begingroup
\setlength{\tabcolsep}{3pt} 
\renewcommand{\arraystretch}{1.05} 
\begin{tabular}{@{}c|ccc|ccc|ccc|ccc|ccc|ccc@{}}
\toprule
\textbf{\# of} 
  & \multicolumn{3}{c|}{\textbf{DePlot}} 
  & \multicolumn{3}{c|}{\textbf{ChartGemma}} 
  & \multicolumn{3}{c|}{\textbf{TinyChart}}
  & \multicolumn{3}{c|}{\textbf{Pixtral}}
  & \multicolumn{3}{c|}{\textbf{Gemini}}
  & \multicolumn{3}{c}{\textbf{GPT-4o}}\\
  \textbf{Entities} 
 & F1 $\uparrow$ & Sig $\uparrow$ & SES $\downarrow$
 & F1 $\uparrow$ & Sig $\uparrow$ & SES $\downarrow$
 & F1 $\uparrow$ & Sig $\uparrow$ & SES $\downarrow$
 & F1 $\uparrow$ & Sig $\uparrow$ & SES $\downarrow$
 & F1 $\uparrow$ & Sig $\uparrow$ & SES $\downarrow$
 & F1 $\uparrow$ & Sig $\uparrow$ & SES $\downarrow$\\
\midrule
1–2 & 47.78 & 50.00 & -1.03 
    & 27.38 & 34.55 & 3.38
    & 65.14 & 69.90 & 0.05
    & 73.57 & 85.23 & -0.44 
    & 87.58 & 97.27 & 1.06
    & 11.25 & 21.34 & 3.84\\
3–4 & 47.59 & 50.27 & -2.33 
    & 18.26 & 24.41 & 7.18
    & 71.62 & 77.87 & -0.14 
    & 67.03 & 81.26 & -2.74 
    & 81.23 & 94.26 & -0.62
    & 7.78 & 15.07 & 7.13\\
5–6 & 34.10 & 36.96 & 3.10
    & 9.57 & 14.32 & 11.42
    & 61.37 & 68.56 & 4.06
    & 59.38 & 75.57 & -3.96
    & 66.69 & 81.56 & 0.67
    & 5.08 & 9.94 & 9.60\\
\bottomrule
\end{tabular}
\caption{Performance by entity count. All metric values are reported as percentages; F1 denotes RMS$^{TBE}_{F1}$; Sig. denotes RMS$^{TBE}_{F1}$-Sig.}
\label{entity}
\endgroup
\end{table*}

\noindent{\bf Biases by the number of major ticks (measured by RMS$^{TBE}_{F1}$):} Our FairChart2Table Part B was used to evaluate the effect of the number of major ticks of 3 and 11. Recall that the base configuration is from Part A, with 6 major ticks. Figure~\ref{spider}(a) and Table~\ref{everything} demonstrate interesting patterns. All the models perform the worst with 3 major ticks. In contrast, for the 11 major ticks, the performance of the closed-source models improves compared to the base configuration, but there is no uniform trend for the open-source models.

\noindent{\bf Biases by the range of y-axis values (measured by RMS$^{TBE}_{F1}$):} We used Part C of our dataset for the experiments. As shown in Figure~\ref{spider}(b) and Table~\ref{everything},
DePlot’s performance slightly increases when the intersection of the y-axis with the x-axis shifts from zero to a positive value, decreases when it moves from zero to a negative value, and is lowest for charts with extended ranges. TinyChart performs the best when the y-axis crosses the x-axis at zero, outperforming DePlot under this base configuration. However, it performs significantly worse than DePlot on the chart, with a displaced y-axis origin. Additionally, TinyChart handles charts with extended ranges better than it does charts with a displaced y-axis origin. ChartGemma’s performance drops substantially when the y-axis origin shifts from zero to a positive value, decreases slightly further when it shifts to a negative value, and is lowest for charts with extended ranges. 
Pixtral performance decreases from zero to positive and then to negative y-axis displacement. In contrast, Pixtral outperforms DePlot at the extended ranges.
Gemini performs consistently well regardless of the y-axis origin's displacement. Its performance is slightly better with a positive displacement and slightly worse with a negative displacement compared to no displacement. In contrast, a huge drop is observed for charts with extended ranges.
On the contrary, GPT-4o performs better for charts with a displaced y-axis origin, compared to charts with the y-axis at the origin, while it performs worse when the range is extended.

\noindent{\bf Biases by the number format on the y-axis:} The dataset Part D was used for evaluation. Figure~\ref{spider}(c) shows the results.
Interestingly, DePlot performs better with the comma format than with the plain numerical format, while its performance significantly declines with the scientific notation and abbreviation formats. ChartGemma performs slightly better with the comma format than with the numerical format, but its performance drops sharply with the abbreviation format and reaches its lowest level with the scientific notation format. TinyChart performs slightly better with the scientific notation format than with the plain numerical format, while its performance significantly declines with the comma and abbreviation formats. Pixtral performs slightly better with the scientific format than with the plain numerical format. While its performance also declines with the comma and abbreviation formats, the drop is moderate for the comma format but dramatic with the abbreviation format.
Gemini performs slightly better with the scientific notation format than with the plain numerical format, while its performance slightly decreases with the comma format. In contrast, there is a dramatic performance drop with the abbreviation format. Table~\ref{everything} shows that GPT-4o performs better with the comma and scientific notation formats than with the plain numerical format, while its performance slightly decreases with the abbreviation format.

\begin{table}[ht]
\small
\centering
\begin{tabular}{c|c}
\toprule
\textbf{Entities} & \textbf{Average Number of Crossing Points} \\
\toprule
2 & 1.450 \\
3 & 3.067 \\
4 & 4.175 \\
5 & 6.120 \\
6 & 7.467 \\
\bottomrule
\end{tabular}
\caption{Average number of crossing points per entity, grouped by number of entities from Part A (different digit length) of FairChart2Table.}
\label{crossing}
\end{table}

\subsection{RQ2: What other factors of chart images impact MLM performance? \label{RQ2}}

\noindent{\bf Biases by the number of entities:}
Figure~\ref{spider}(d) shows a general trend that the model performance generally decreases as the number of entities increases, especially DePlot, ChartGemma, and Pixtral. In contrast, Gemini and TinyChart are more robust, although TinyChart shows a bigger performance drop at 6 entities compared to Gemini.
As the number of entities increases, models are more likely to confuse numerical values on the same x-axis because of the higher chance of crossing points. We did not explicitly control the number of line crossings when varying the number of entities. However, Table~\ref{crossing} shows the high correlation between the number of entities and the average number of crossing points per entity with the Pearson correlation of 0.9549. This strong correlation suggests that, in our setup, line crossings are effectively mediated by the entity count, allowing SES to be a meaningful metric for this evaluation. Low SES scores are desirable. As shown in Table~\ref{entity}, DePlot has a high SES score with 5 and 6 entities, TinyChart and GPT-4o also show a dramatic increase with 6 entities, and ChartGemma shows a gradual increase as the number of entities increases. In contrast, Pixtral's SES score decreases as the number of entities increases, and Gemini shows only a modest increase.

\begin{table}[ht]
\centering
\scriptsize
\setlength{\tabcolsep}{6.5pt}
\begin{tabular}{@{}c|c|c|c|c|c|c@{}}
\toprule
\textbf{Type} & \textbf{DePlot} & \shortstack{\textbf{Chart}\\\textbf{Gemma}} & \shortstack{\textbf{Tiny}\\\textbf{Chart}} & \textbf{Pixtral} & \textbf{Gemini} & \textbf{GPT-4o}\\
\midrule
Line & 41.64 & 12.97 & 65.12 & 55.81 & 72.01 & 8.39\\
Dot  & 46.13 & 12.70 & 67.91 & 72.30 & 81.87 & 6.93\\
Bar  & 41.69 & 29.53 & 65.09 & 71.85 & 81.62 & 8.76\\
\bottomrule
\end{tabular}
\caption{RMS$^{TBE}_{F1}$ across different chart types.}
\label{types}
\end{table}

\noindent{\bf RMS$^{TBE}_{F1}$-Sig by Entity Count:}
As shown in Table~\ref{entity}, the gap between RMS$^{TBE}_{F1}$-Sig and RMS$^{TBE}_{F1}$ increases for all models except for ChartGemma and GPT-4o, indicating that the increasing prediction errors are still within the minor tick threshold. In contrast, for ChartGemma and GPT-4o, the gap decreases, implying more prediction errors fall outside the threshold. Both metric values fall, and the gap narrows.

\noindent{\bf Impact of chart types:} MLM performance appears to vary across chart types, as reported by \citet{simplot}. However, as shown in Table~\ref{types}, when the underlying data for each chart type are the same, the model performance differences become less pronounced but still statistically significant except for three cases shown in Appendix~\ref{type_test}. 

\subsection{RQ3: Does prompting MLMs with explicit y-axis information help improve MLM performance?  \label{RQ3}}

For general-purpose MLMs, we introduce a y-axis hinting method that enables the model to perform chart-to-table translation conditioned on manually extracted y-axis tick values. See the prompts in Appendix~\ref{prompt}.
Table~\ref{everything} illustrates that Pixtral exhibits substantial gains using our y-axis tick-value hints, ranging from 0.94 to 26.60. Moreover, the prompting strategy also enhances GPT-4o's performance, except for the comma and scientific notation formats. In contrast, Gemini performs generally similarly or only slightly better with the additional information, but when labels are abbreviated, its performance increases dramatically, with the score rising from 46.54 to 61.32.

\section{Conclusions and Future Work}

We propose the FairChart2Table benchmark and demonstrate biases related to the y-axis information for chart-to-table translation. Our benchmark can be used to measure similar biases in other MLMs. Using manually extracted y-axis tick values as prompt hints can significantly reduce the bias related to digit lengths. Future work includes automated debiasing methods for chart-to-table translation tasks.


\section{Limitations}

This paper provides a comprehensive evaluation of how y-axis information affects MLM performance in chart-to-table translation. It has the following limitations. First, the proposed dataset consists of three chart types and does not include real-world chart images. However, to study bias associated with one aspect, such as digit length, we need to keep the other aspects fixed and balanced to prevent the confounding impact from the other aspects. Synthetic data generation is required. Second, charts and legends are only in English, as in existing research~\citep{deplot}. 
Lastly, this study does not focus on biases unrelated to the y-axis chart-to-table translation. Our FairChart2Table dataset and framework can be extended to include other chart types and non-English headers.

\section{Ethical Considerations and Potential Risk}

The authors follow the ACL Code of Ethics. The ideas presented in this paper regarding investigating y-axis-related biases, experimental designs, and conclusions are solely those of the authors. AI assistants were utilized only for writing refinement, presentation, and basic coding support, including formatting and debugging. Since the dataset is synthetically generated, it is not subject to privacy concerns. All scripts, evaluation metrics, datasets, and prompting methods will be made publicly available for reproducibility. Our findings show that the degree of bias varies across different understudied MLMs. Therefore, specific findings may not generalize to other MLMs or to the same models under different training or fine-tuning conditions. Nevertheless, our FairChart2Table framework can be used to investigate the y-axis biases in other MLMs in a similar way.

\section{Acknowledgement}

This work is partially supported by the National Science Foundation under Grant No. 2152117. Any opinions, findings, and conclusions or recommendations expressed in this material are those of the author(s) and do not necessarily reflect the views of the National Science Foundation.

\bibliography{custom}

\appendix

\section{Appendix}
\label{sec:appendix}

\subsection{Details about RMS$^{TBE}_{F1}$}
\label{sec:appendix_rms}

The similarity score with the distance between two numerical values is defined as:

\begin{equation} D_{\text{TBE}}(g_j, p_i) = min \left(  1, \frac{\lvert g_j - p_i \rvert}{t} \right), \end{equation}
where $\mathbf{g} = \{g_j\}_{1\leq j\leq m}$ for the ground truth column with $m$ elements and $\textbf{p} = \{p_i\}_{1\leq i\leq n}$ for the predicted column with $n$ elements, respectively. In our implementation, the value of $t$ is one-fifth of the interval between consecutive major ticks of a given chart image. Following the minimal cost matching procedure used in RMS~\citep{deplot}, we align the predicted and ground-truth datapoints using only \textit{Normalized Levenshtein Distance} between their row and column headers. Let $X \in \{0,1\}^{n \times m}$ denote the resulting binary assignment matrix, where $X_{ij}=1$ if the predicted header of datapoint $i$ is matched to the ground-truth header of datapoint $j$, and $X_{ij}=0$ otherwise. Once the assignment is fixed, we compute precision and recall using only $D_{\text{TBE}}(g_j, p_i)$  for the matched value pairs, without incorporating header similarity into the final value-level score.

\begin{equation}
S_{\mathrm{TBE}}(g_j, p_i)=1-D_{\mathrm{TBE}}(g_j, p_i)
\end{equation}

\begin{equation} \text{RMS}^{TBE}_{precision} = \frac{\Sigma^n_i\Sigma^m_j X_{ij}S_{\text{TBE}}(g_j, p_i)}{n} \end{equation}

\begin{equation} \text{RMS}^{TBE}_{recall} = \frac{\Sigma^n_i\Sigma^m_j X_{ij}S_{\text{TBE}}(g_j, p_i)}{m} \end{equation}

RMS$^{TBE}_{F1}$ is the harmonic mean of the above precision and recall.

\subsection{Results of RMS$_{F1}$ with and without header similarity scores \label{app:rms-results}} 

\begin{figure*}[h!tb] {
\centering
\includegraphics[width=0.9\textwidth]{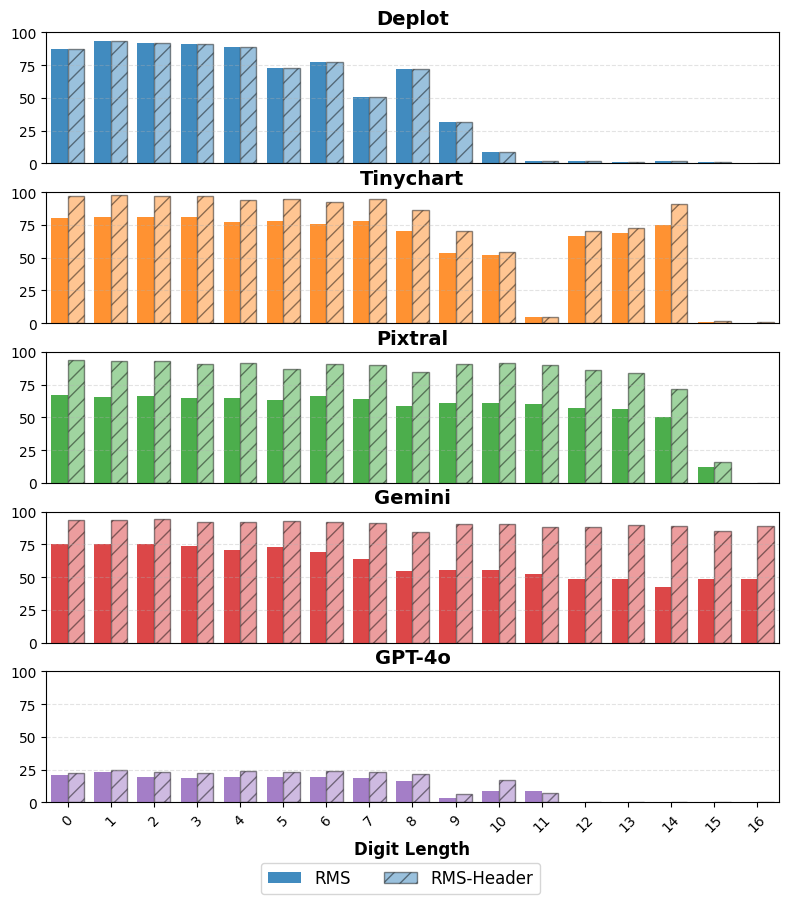} 
\caption{Bar charts comparing RMS$_{F1}$ and RMS-Header (RMS$_{F1}$ without header scores) by different models.}
\label{rms_and_header}
}
\end{figure*}

As Figures~\ref{rms_and_header} illustrates, some models are over-penalized a lot because of headers. We compare RMS$_{F1}$ and RMS$_{F1}$ without header score on FairChart2Table data. As the figure illustrates, different models generate distinct outputs for various headers, and that affects the accuracy.



\subsection{Result of Statistical Test for Impact of Chart Types on MLM performance \label{type_test}}

See Table~\ref{type_test_table}.

\subsection{Prompts for Gemini and Pixtral \label{prompt}}

See Tables~\ref{prompt1} - \ref{prompt2}.

\subsection{Examples FairChart2Table Chart Images \label{app:image}}

\subsubsection{Part A}

\begin{figure*}[h!tb] {
\centering
\includegraphics[width=0.7\textwidth]{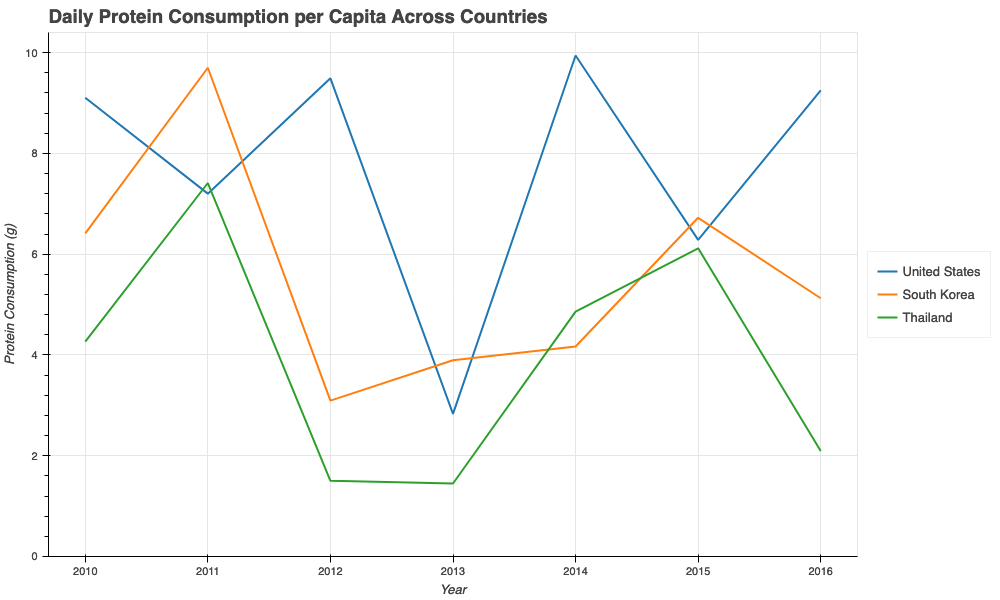} 
\caption{Part A: Line Chart}
\label{base_1}
}
\end{figure*}

\begin{figure*}[h!tb] {
\centering
\includegraphics[width=0.7\textwidth]{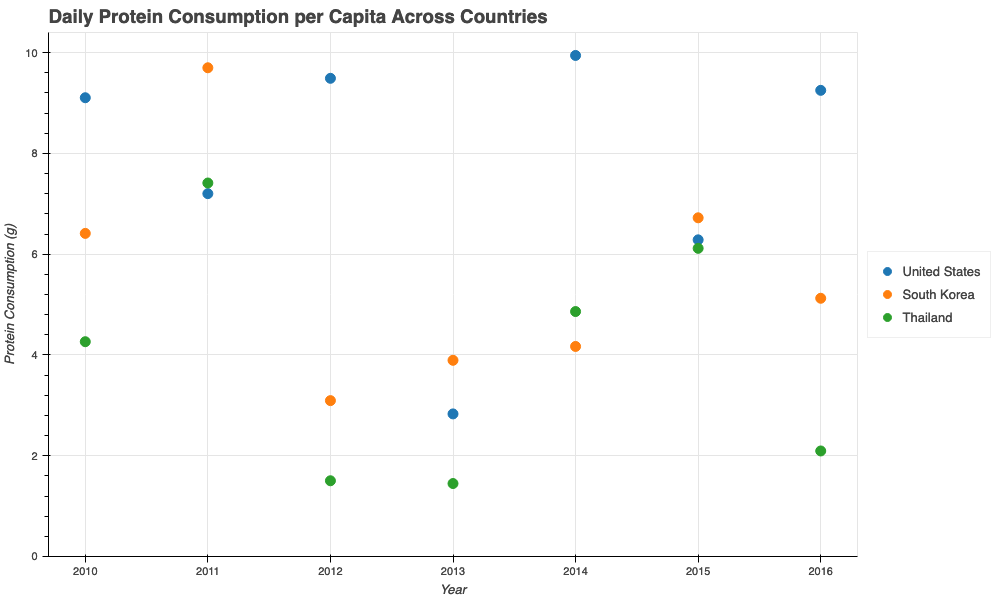} 
\caption{Part A: Dot Chart}
\label{base_2}
}
\end{figure*}

\begin{figure*}[h!tb] {
\centering
\includegraphics[width=0.7\textwidth]{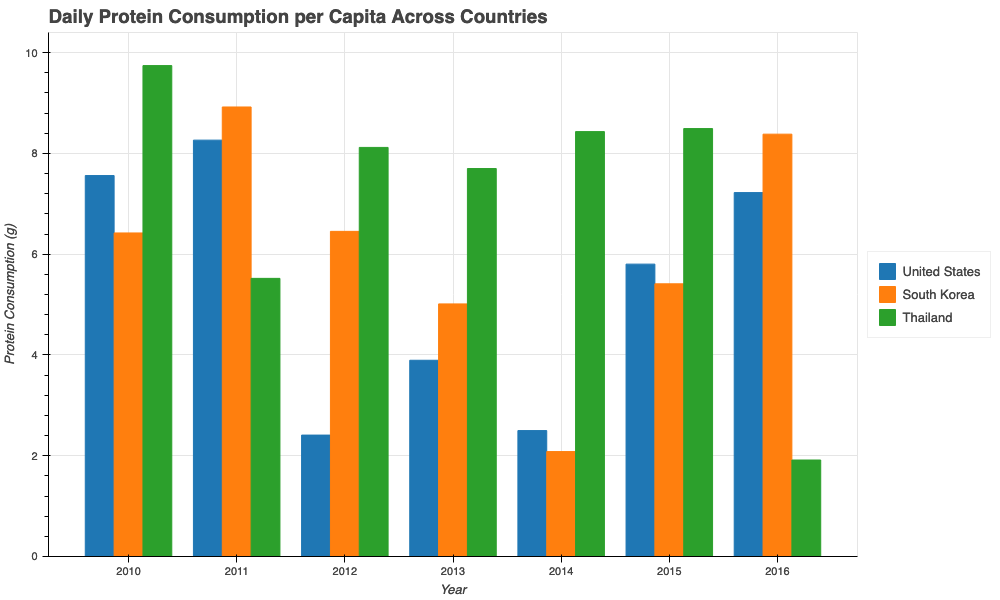} 
\caption{Part A: Bar Chart}
\label{base_3}
}
\end{figure*}

\begin{figure*}[h!tb] {
\centering
\includegraphics[width=0.7\textwidth]{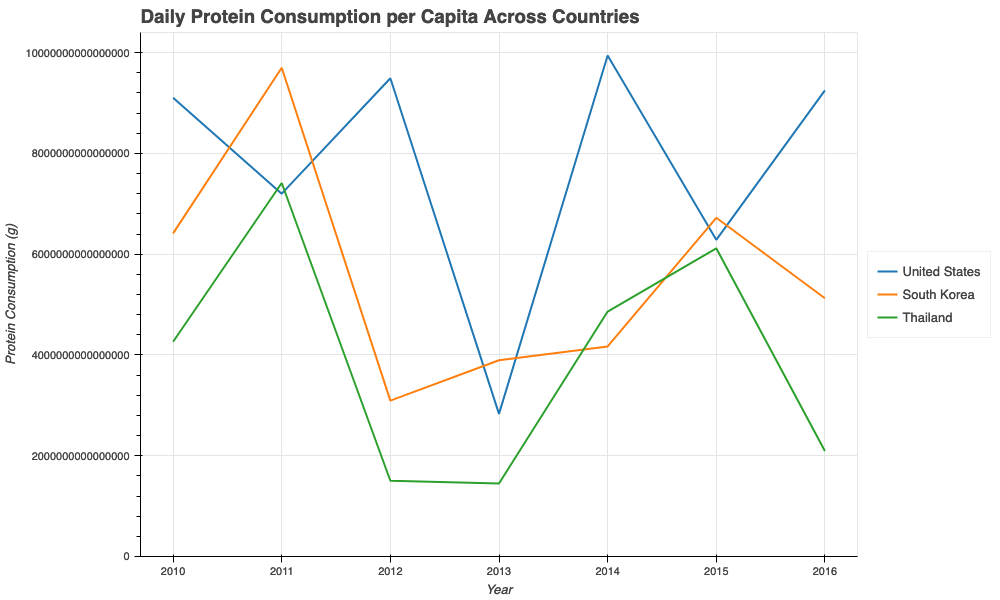} 
\caption{Part A: Line Chart at Digit Length 16}
\label{base_4}
}
\end{figure*}

See Figures~\ref{base_1} - \ref{base_4}.
\subsubsection{Part B}

\begin{figure*}[h!tb] {
\centering
\includegraphics[width=0.7\textwidth]{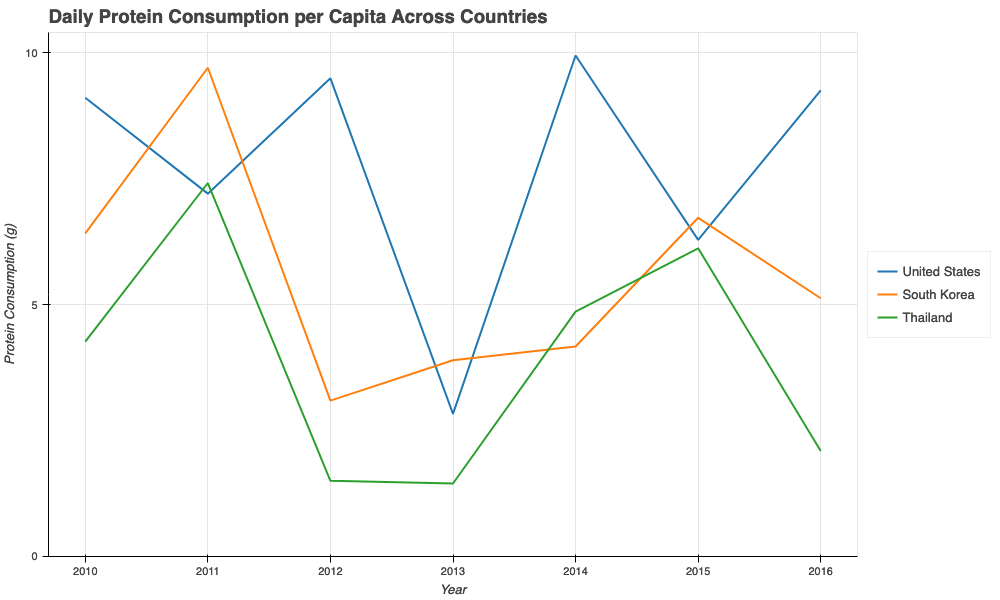} 
\caption{Part B:  Line Chart at Digit Length 1 with 3 Major Ticks}
\label{major_3}
}
\end{figure*}

See Figures~\ref{major_3} - \ref{major_11}.

\subsubsection{Part C}

\begin{figure*}[h!tb] {
\centering
\includegraphics[width=0.7\textwidth]{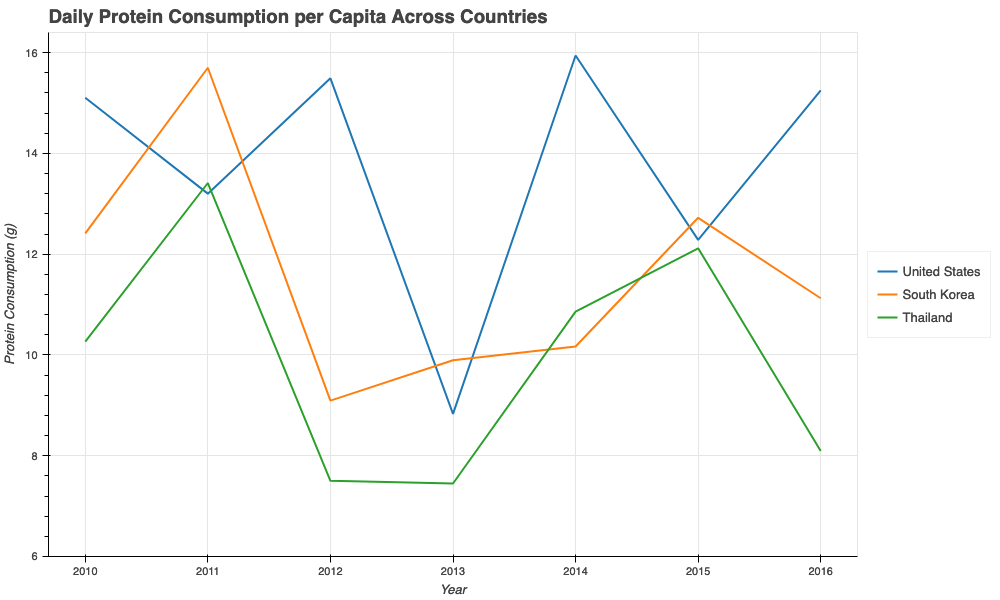} 
\caption{Part C: Line Chart with Positive Minimum Tick Value transformed from Digit Length 1}
\label{pos}
}
\end{figure*}

\begin{figure*}[h!tb] {
\centering
\includegraphics[width=0.7\textwidth]{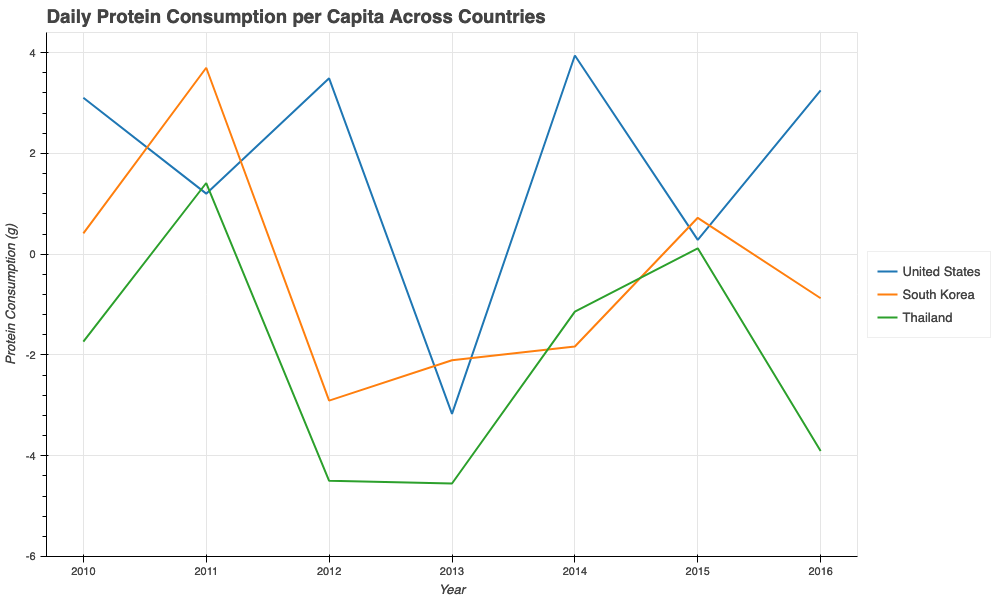} 
\caption{Part C: Line Chart with Negative Minimum Tick Value transformed from Digit Length 1 }
\label{neg}
}
\end{figure*}

\begin{figure*}[h!tb] {
\centering
\includegraphics[width=0.7\textwidth]{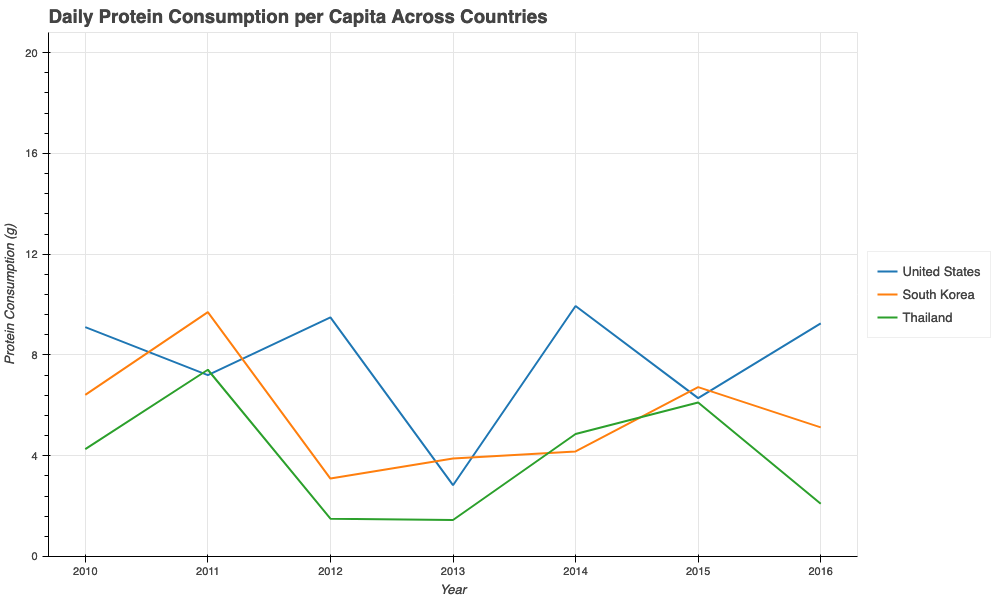} 
\caption{Part C: Line Chart at Digit Length 1 with Extended Range}
\label{squ}
}
\end{figure*}

See Figures~\ref{pos} - \ref{squ}.

\subsubsection{Part D}

\begin{figure*}[h!tb] {
\centering
\includegraphics[width=0.7\textwidth]{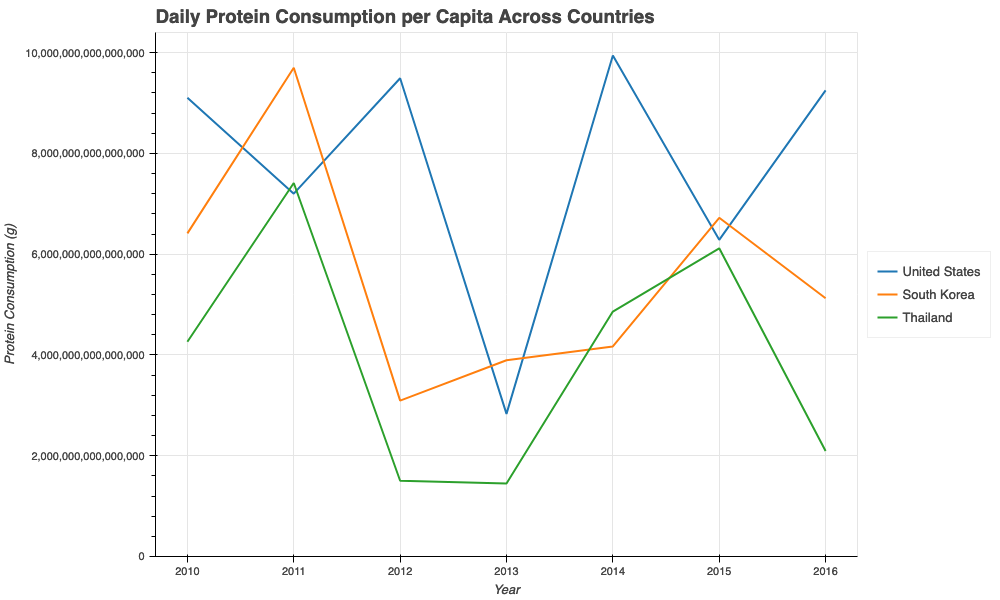} 
\caption{Part D: Line Chart at Digit Length 16 with Comma Format}
\label{comma}
}
\end{figure*}

\begin{figure*}[h!tb] {
\centering
\includegraphics[width=0.7\textwidth]{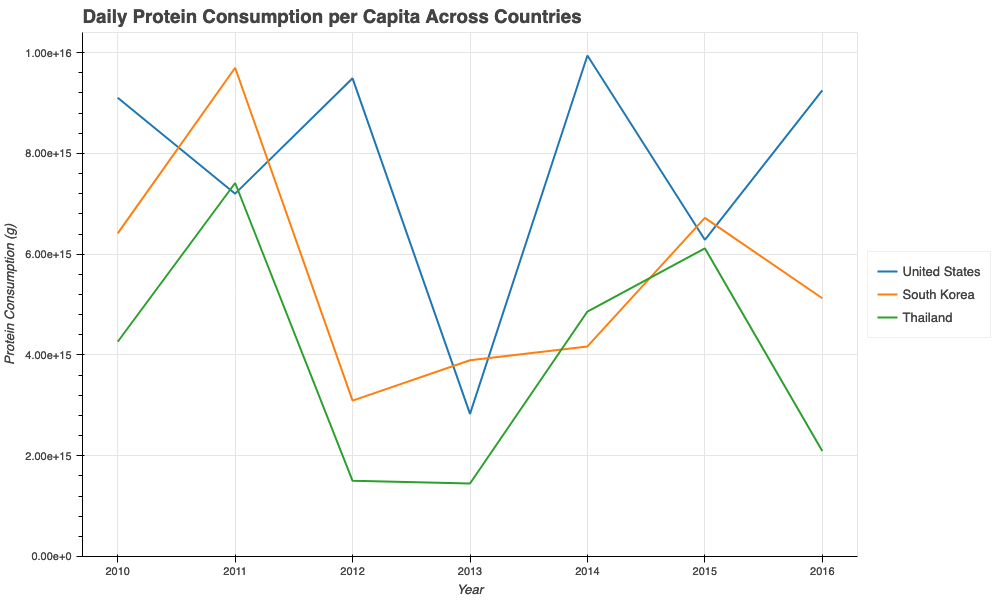} 
\caption{Part D: Line Chart at Digit Length 16 with Scientific Notation Format}
\label{sci}
}
\end{figure*}

\begin{figure*}[h!tb] {
\centering
\includegraphics[width=0.7\textwidth]{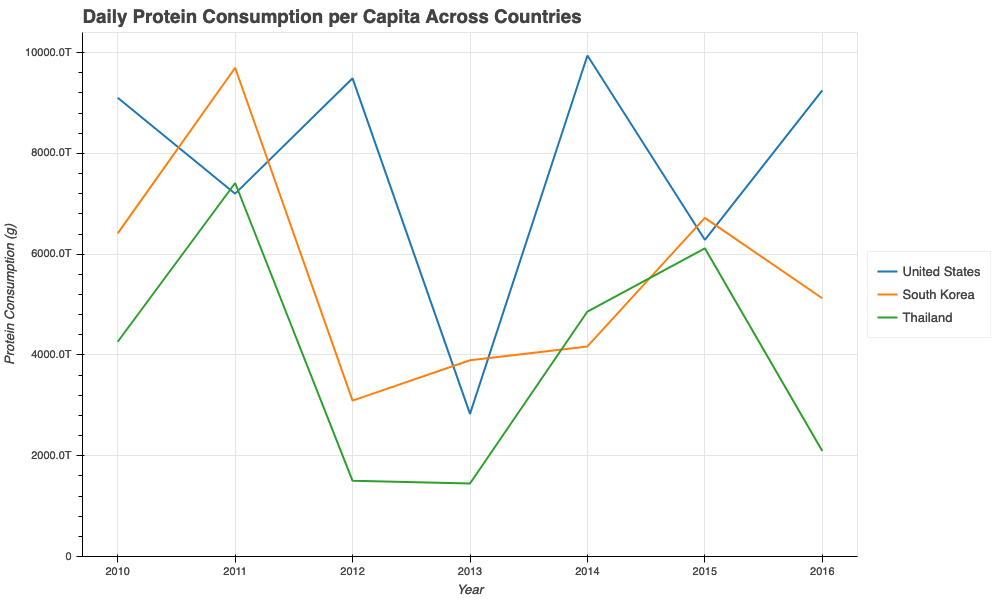} 
\caption{Part D: Line Chart at Digit Length 16 with Abbreviation Format}
\label{abb}
}
\end{figure*}

See Figures~\ref{comma} - \ref{abb}.

\subsection{Examples of Original Chart and Generated Chart Based on Incorrectly Predicted Tables By DePlot}

\begin{figure*}[h!tb] {
\centering
\includegraphics[width=0.95\textwidth]{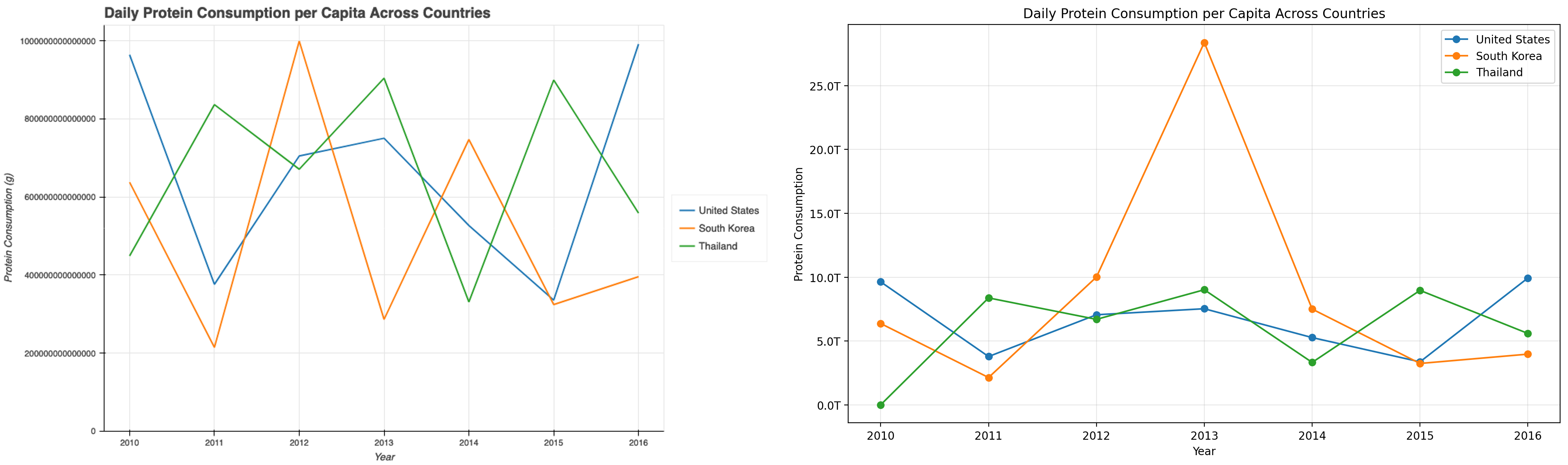} 
\caption{Comparison between the original chart (left) and the chart predicted by DePlot (right). The predicted chart reflects a long-digit-length bias, returning values with incorrect digit lengths.}
\label{side_digit}
}
\end{figure*}

\begin{figure*}[h!tb] {
\centering
\includegraphics[width=0.95\textwidth]{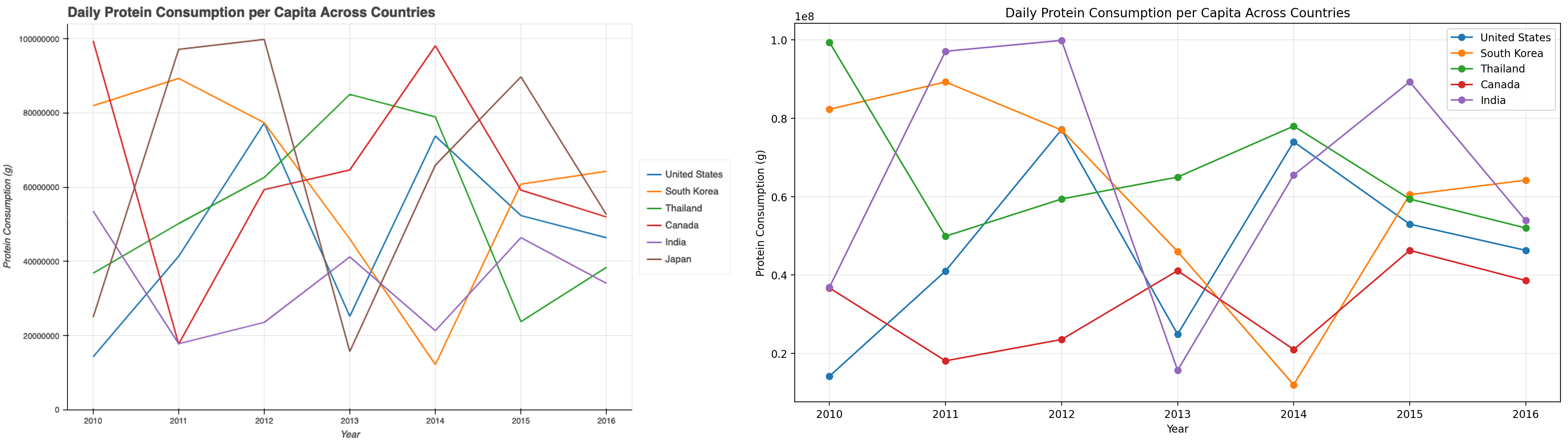} 
\caption{Comparison between the original chart (left) and the chart predicted by DePlot (right). The predicted chart shows confusion caused by many entities and frequent line crossings, leading to mismatched values across categories.}
\label{side_cross}
}
\end{figure*}

\begin{figure*}[h!tb] {
\centering
\includegraphics[width=0.95\textwidth]{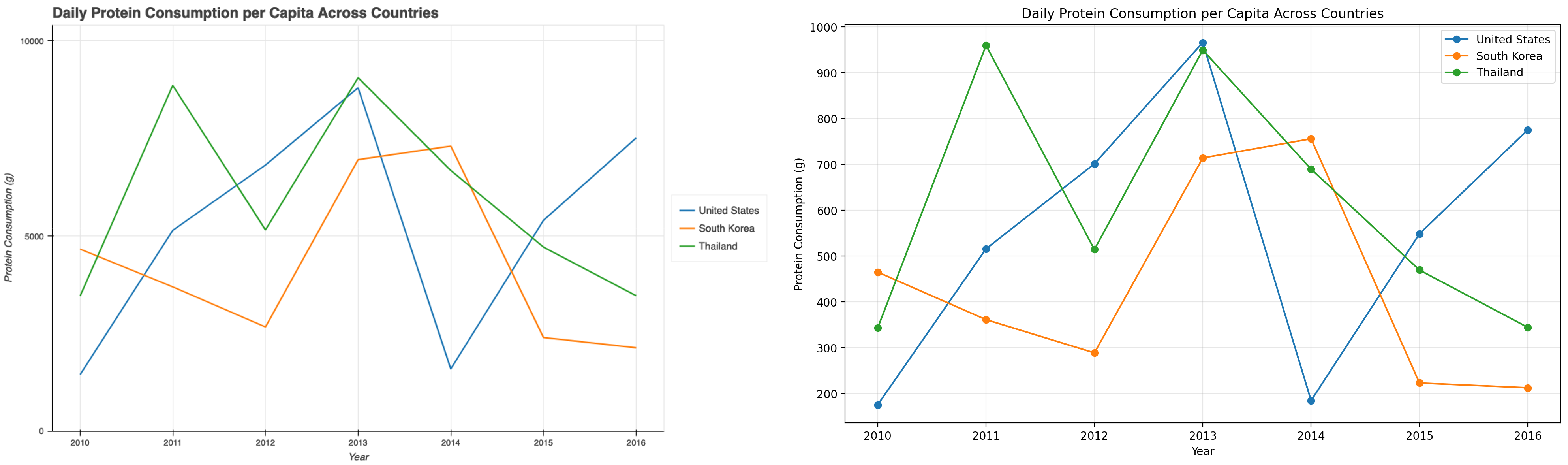} 
\caption{Comparison between the original chart (left) and the chart predicted by DePlot (right). With only three major ticks, the predicted chart preserves a similar overall shape but distorts the proportions.}
\label{side_3}
}
\end{figure*}

\begin{figure*}[h!tb] {
\centering
\includegraphics[width=0.95\textwidth]{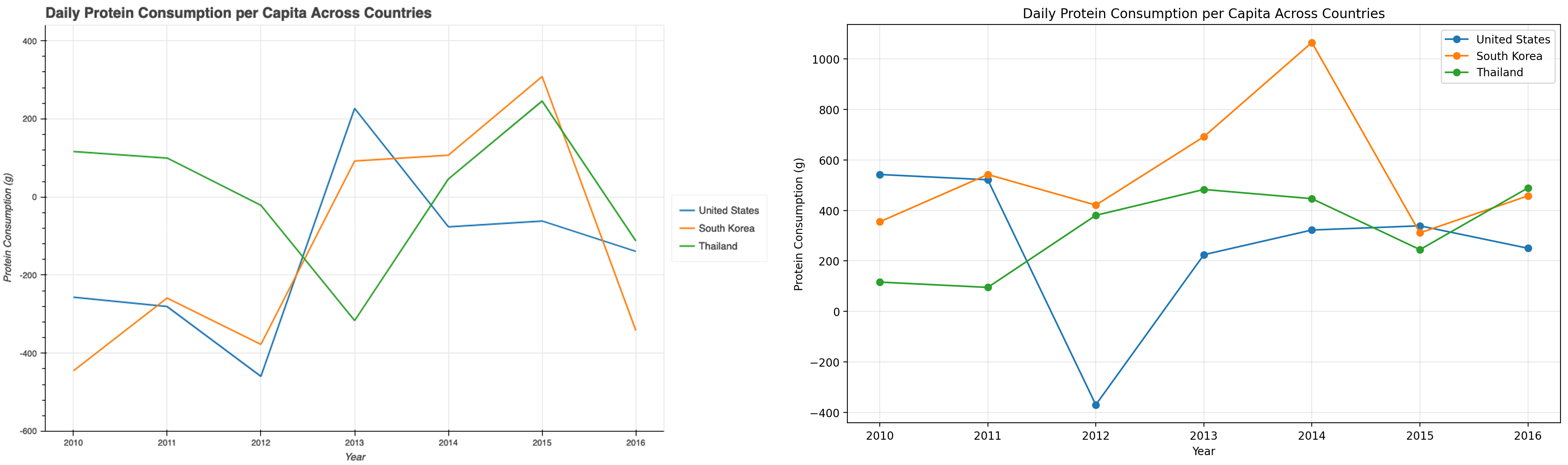} 
\caption{Comparison between the original chart (left) and the chart predicted by DePlot (right). When the y-axis origin is negative, the predicted chart misestimates the point values.}
\label{side_neg}
}
\end{figure*}

See Figures~\ref{side_digit} - \ref{side_neg}.

\begin{table*}
\small
\centering
\begin{tabular}{c|c|c|c|c|c|c}
\toprule
\textbf{Type} & \textbf{DePlot} & \textbf{ChartGemma} & \textbf{Tiny} & \textbf{Pixtral} & \textbf{Gemini} &\textbf{GPT}\\
\toprule
Dot VS. Bar  & 5.067968e-02 & \cellcolor{yellow}1.85467e-67 & \cellcolor{yellow}4.041066e-02 & \cellcolor{yellow}1.318864e-02 & \cellcolor{yellow}2.032969e-08 & \cellcolor{yellow}5.515060e-18\\
Line VS. Bar  & \cellcolor{yellow}7.570325e-14 & \cellcolor{yellow}5.44866e-72 & \cellcolor{yellow}2.044370e-28 & \cellcolor{yellow}1.147831e-108 & \cellcolor{yellow}8.373813e-126 & 1.269979e-01 \\
Line VS. Dot  & \cellcolor{yellow}1.419833e-07 & 0.871489 & \cellcolor{yellow}5.968554e-19 & \cellcolor{yellow}4.882375e-110 & \cellcolor{yellow}6.484573e-118 & \cellcolor{yellow}2.871905e-12 \\
\bottomrule
\end{tabular}
\caption{P-values of the Wilcoxon Signed-rank test. Except for Dot vs. Bar with Deplot and Line vs. Bar with GPT, all p-values are smaller than 0.05. \colorbox{yellow}{Yellow} indicates that the comparison is statistically significant.}
\label{type_test_table}
\end{table*}

\begin{figure*}[h!tb] {
\centering
\includegraphics[width=0.7\textwidth]{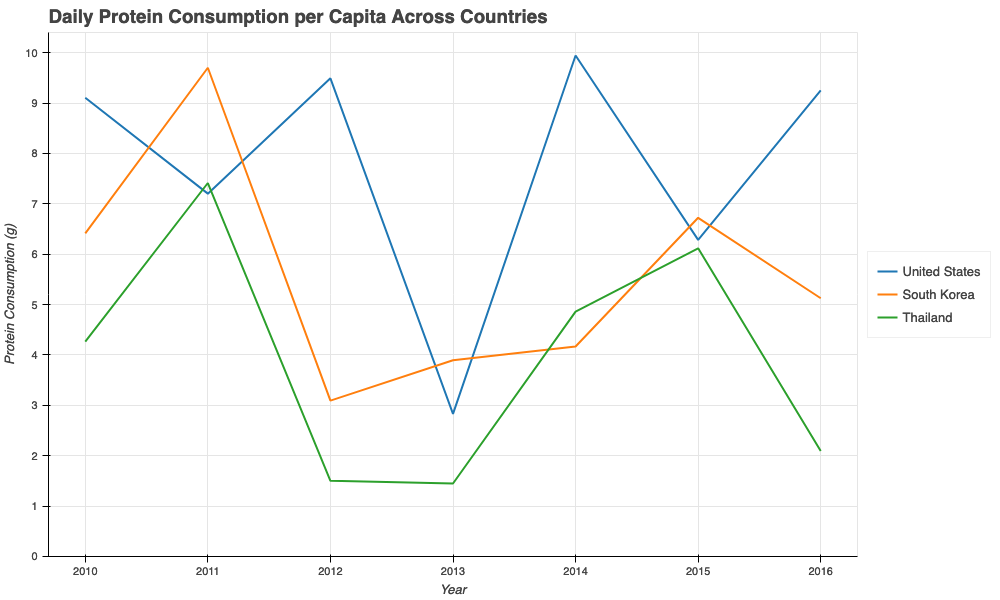} 
\caption{Part B: Line Chart at Digit Length 1 with 11 Major Ticks}
\label{major_11}
}
\end{figure*}

\begin{table*}[!htbp]
\centering
\begin{tabular}{l}
\hline
\textbf{Prompt: }\\
Generate underlying data table for the chart.
\\
\hline
\end{tabular}
\caption{Prompt for chart-to-table translation}
\label{prompt1}
\end{table*}

\begin{table*}[!htbp]
\centering
\begin{tabular}{l}
\hline
\textbf{Prompt: }\\
Generate underlying data table for the chart. Hint: y-axis major ticks are ...
\\
\hline
\end{tabular}
\caption{Prompt for our hint strategy using major tick values in scientific notation for chart-to-table translation}
\label{prompt2}
\end{table*}

\begin{figure*}[h!tb] {
\centering
\includegraphics[width=0.7\textwidth]{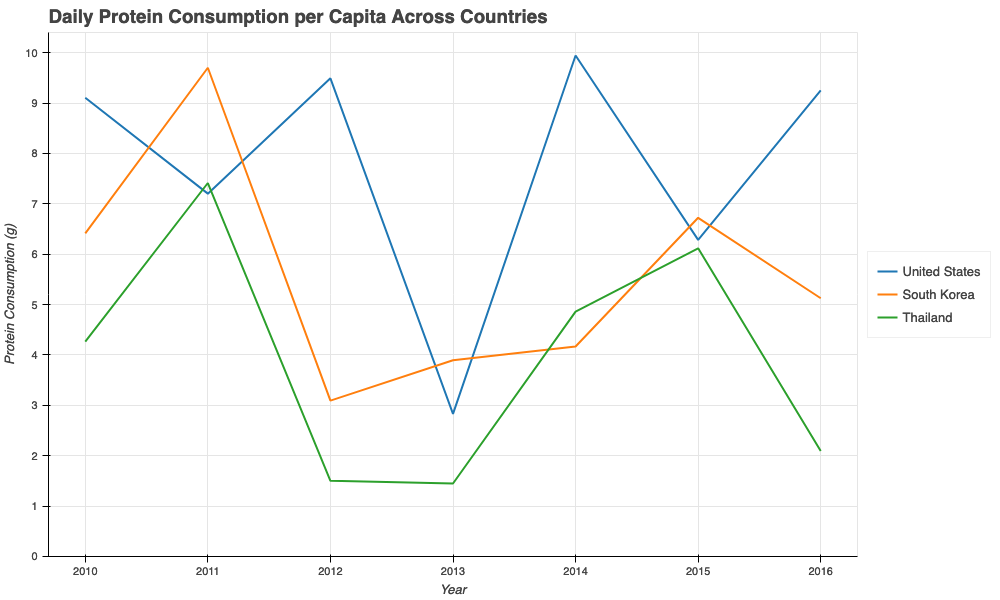} 
\caption{Part B: Line Chart at Digit Length 1 with 11 Major Ticks}
\label{major_11}
}
\end{figure*}

\end{document}